# A Numerical Optimization Algorithm Inspired by the Strawberry Plant[1]

F. Merrikh-Bayat

Department of electrical and computer engineering, University of Zanjan, Zanjan, Iran.

Tel.: +98 24 3305 4061, E-mail address: f.bayat@znu.ac.ir

**Abstract:** This paper proposes a new numerical optimization algorithm inspired by the strawberry plant for solving complicated engineering problems. Plants like strawberry develop both runners and roots for propagation and search for water resources and minerals. In these plants, runners and roots can be thought of as tools for global and local searches, respectively. The proposed algorithm has three main differences with the trivial nature-inspired optimization algorithms: duplication-elimination of the computational agents at all iterations, subjecting all agents to both small and large movements from the beginning to end, and the lack of communication (information exchange) between agents. Moreover, it has the advantage of using only three parameters to be tuned by user. This algorithm is applied to standard test functions and the results are compared with GA and PSO. The proposed algorithm is also used to solve an open problem in the field of robust control theory. These simulations show that the proposed algorithm can very effectively solve complicated optimization problems.

---

[1] Please be advised that this is the draft of a paper submitted to a journal for possible publication. The final version may have differences with this one according to the changes made during the review process.

## 1. Introduction

During the past four decades nature had been the source of inspiration for developing new optimization algorithms for solving complicated engineering problems. The first attempts to find a cybernetic solution to a practical problem can be found in the works of Rechenberg [1]. However, the first general-purpose and well-explained algorithm of this type is probably the genetic algorithm (GA) developed by Holland [2]. At this time, various nature-inspired optimization algorithms are available, among them the GA, Particle Swarm Optimization (PSO) [3,4], Ant Colony Optimization (ACO) [5,6], Artificial Bee Colony (ABC) algorithm [7,8], Simulated Annealing (SA) [9,10], Firefly Algorithm (FA) [11], Bacterial Foraging Optimization (BFO) [12], Artificial Immune System (AIS) [13,14], shuffled frog-leaping algorithm [15], Differential Evolution (DE) [16,17], and Imperialist Competitive Algorithm (ICA) [18], have attracted more attentions. It is a well-known fact that meta-heuristic optimization algorithms are capable of solving many complicated engineering problems whose solutions cannot effectively be obtained by using classical (often, gradient-based) optimization algorithms. Numerous successful applications of these algorithms can be found in the literature (see for example [19]-[21]).

Although the source of inspiration is different in nature-inspired optimization algorithms, they still have many similarities. The main relationships between various nature-inspired optimization algorithms are: application of random variables, ability of dealing with uncertain and non-differentiable cost functions, simultaneous application of more than one computational agent for searching the domain of problem (except SA which applies only one agent), existing a kind of communication scheme between computational agents (e.g., the crossover operator in GA, the social term in PSO, pheromone trail in ACO, dancing of artificial bees in ABC, the light emission in FA, etc.), application of the objective function itself rather than its derivative for performing the search, and elimination of weak solutions at every iteration. In brief, it can be said that all of these algorithms perform a kid of *memory stochastic search,* which aims to optimize a certain objective function. More precisely, all algorithms of this type model the behavior of a (colony of) certain living thing or a certain physical phenomena by performing a kind of optimization. The procedure of modeling is always iterative and makes use of random variables, but the method is not wholly stochastic and has a kind of memory to remember the good solutions of previous iterations and provide the fittest agents of the colony with a more chance to survive and reproduce.

Clearly, every meta-heuristic optimization algorithm has its own advantages and disadvantages, and actually, it is pointless to look for the *best algorithm* [22]. For example, any algorithm of this type has some *tuning parameters*, which are often adjusted by trial and error. As a general rule, smaller the number of tuning parameters, more advantageous the algorithm is. Using this tradition, the Intelligent Water Drops (IWD) algorithm [23] is not so favorite since it applies too many parameters to be tuned without providing a rigorous method for this purpose. PSO is fast but sometimes (in the standard version) accuracy of solutions is not improved by increasing the number of iterations, and moreover, when it is trapped in a local optimum it has a small chance to escape and continue the search. High sensitivity to the initial guess and low probability of finding the global best solution are the main drawbacks of SA. GA is often slow and the solutions have a limited accuracy (according to the coding issue) but, compared to PSO, it has the advantage that never leads to a solution outside the region defined by the boundary values of variables. A similar discussion goes on other optimization algorithms. In general, algorithms with smaller number of parameters (to be adjusted by user by trial and error), faster convergence and higher probability of skipping from local optimums are identified as more effective algorithms (the accuracy can be improved by performing a complementary search after the original algorithm stops). It is also important to note that the effectiveness of a certain algorithm strictly depends on the problem it is going to solve. In other words, it may happen that a certain algorithm be very successful in dealing with a problem while it is rather unsuccessful in dealing with another one. For this reason researchers apply different algorithms to a certain problem to find the best method suited to solve it.

The aim of this paper is to propose a numerical optimization algorithm inspired by the strawberry plant for solving continuous multi-variable problems. One main difference between the proposed algorithm and other nature-inspired optimization algorithms is that in this algorithm the number of computational agents is not (uniformly) constant from the beginning to end. In fact, at every iteration the number of computational agents is duplicated in an appropriate manner and then half of the weakest agents are subjected to death. The other difference between the proposed algorithm and others is that in our algorithm any computational agent is subjected to both small and large movements repeatedly from the beginning to end, which makes it possible to perform both the local and global searches simultaneously. Moreover, unlike other meta-heuristic algorithms, in the proposed algorithm the computational agents do not communicate with each other, and the above mentioned duplication-elimination procedure combined with a kind of *stochastic sift* motivates the agents toward the global best solution.

The rest of this paper is organized as the following. In Section 2 we briefly review the method of propagation of the strawberry plant, as well as many others, in nature. Based on these results we present the proposed strawberry algorithm (SBA) in Section 3. Numerical simulations are presented in Section 4, and a practical problem is solved using this method in Section 5. Finally, Section 6 concludes the paper.

## 2. Strawberry Plant in Nature

Animals can tolerate environmental changes much better than plants since they have muscle and brain and can decide to move to places with better conditions whenever it is needed. Probably, the most famous examples of this type are the birds which immigrate to warmer places when the weather gets cold. But, the plants are connected to the earth through their roots and cannot move to places with desired conditions anyway. However, some grasses and plants (such as strawberry) can be propagated through the so-called *runner* (or *stolon*) as shown in Fig. 1[2] (vegetative propagation). The runner is a creeping stalk produced in the leaf axils and grows out from the *mother* (parent) plant. At the second node of runner a new plant, called *daughter plant*, is formed and then again a new runner arises on the daughter plant to generate another new daughter (see Fig. 1(a)). Initially, runner plants produce fewer roots but thereafter put forth excessive fibrous roots and when acquired sufficient growth and roots, the daughter plants can be separated from the mother plant and continue their life individually as the new mother plants. Reproduction of strawberry (as well as other similar plants as shown in Fig. 1) can be though of as a kind of *plant movement* since both the mother and daughter plant have exactly the same genes and they are actually a same plant. However, the mother plant commonly dies sooner than the daughter plant provided that the daughter does not arrive at a location with a very bad condition by chance.

---

[2] The authors of Figs. 1(a)-(c) are Jessica Reeder, Arne Hückelheim, and Macleay Grass Man, respectively. Permitted use.

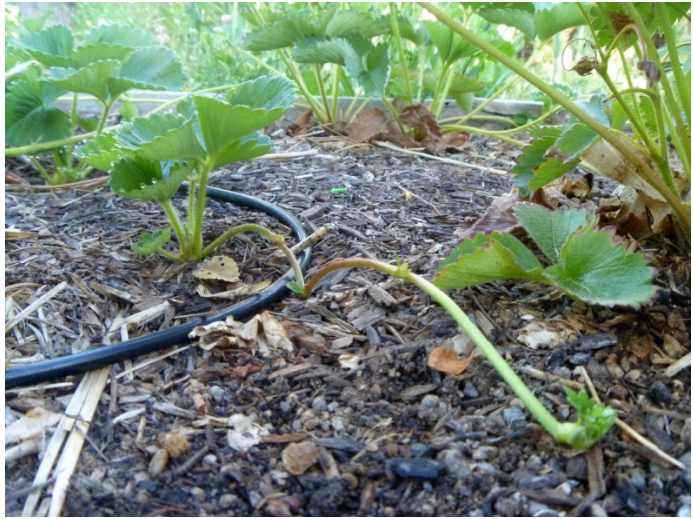

(a)

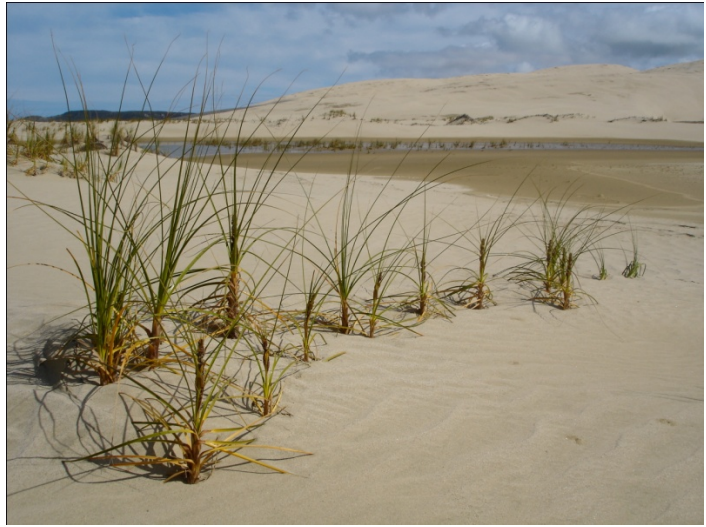

(b)

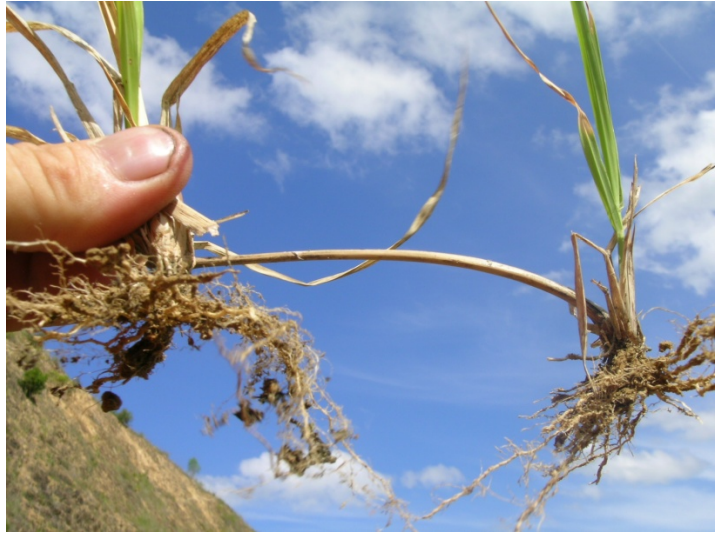

(c)

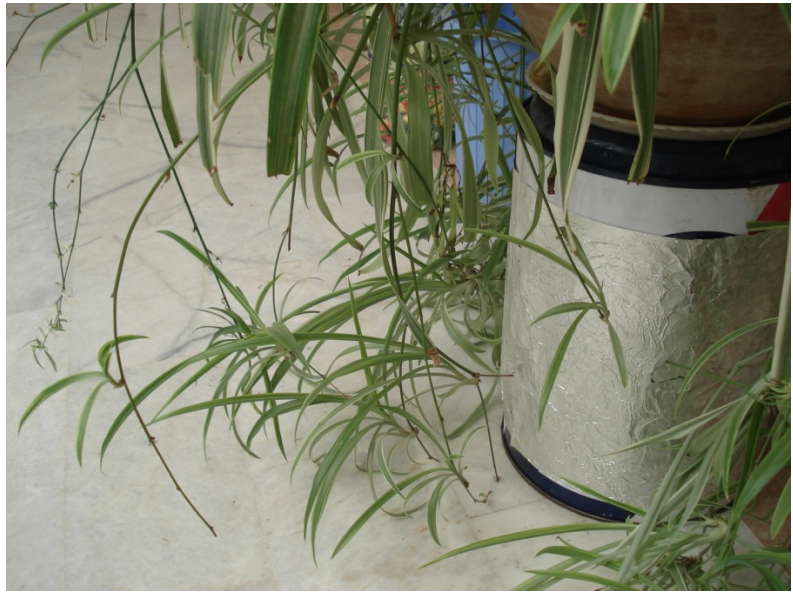

(d)

Fig. 1 (a) Strawberry plant, (b) Desmoschoenus spiralis, (c) Chloris gayana, (d) spider plant. Fig. 1(c) shows the function of runners and roots for global and local search, respectively.

From the mathematical point of view, the plants with runners (such as strawberry) perform a kind of optimization. More precisely, such plants simultaneously perform both the *global* and

*local* search to find resources like water and minerals by developing runners and roots (as well as root hairs), respectively. Both the runners and roots are developed almost randomly, but when a runner or a root (hair) arrives at a place with more water resources, the corresponding daughter plant generates more roots and runners, which affects the growth of the whole plant as well.

Obviously, in order to arrive at a numerical optimization algorithm inspired by the strawberry plant we need to model the behavior of this plant by simple and still explicit rules. In this paper it is assumed that the behavior of strawberry plant can effectively be modeled through the following three facts:

- Each strawberry mother plant is propagated through the runners which rise randomly (global search for resources).
- Each strawberry mother plant develops roots and root hairs randomly (local search for resources).
- Strawberry daughter plants which have access to richer resources grow faster and generate more runners and roots, and on the other hand, the runners and roots which move toward poor resources are more probable to die.

According to the above discussion, in the proposed SBA first we randomly generate certain number of points (computational agents) in the domain of problem, each of them can be though of as a mother plant (it will be shown later that the objective function does not need to be evaluated at the locations of mother plants). Then, at each iteration any mother plant generates one root and one runner (daughter plant); the first one in its vicinity and the other one in a relatively farther location. In fact, in SBA it is assumed that the computational agents consist of runners and roots which move with large and small random steps in the domain of problem, respectively. Next, the objective function is evaluated at the points referred to by runners and roots, and half of these points (which probably have higher fitness values) are selected by, e.g., the roulette wheel or elite selection and considered as the mother plants of next iteration and the remaining half are subjected to death. This procedure is repeated until a predetermined termination condition is satisfied. Figure 2 shows the flowchart of SBA as described above.

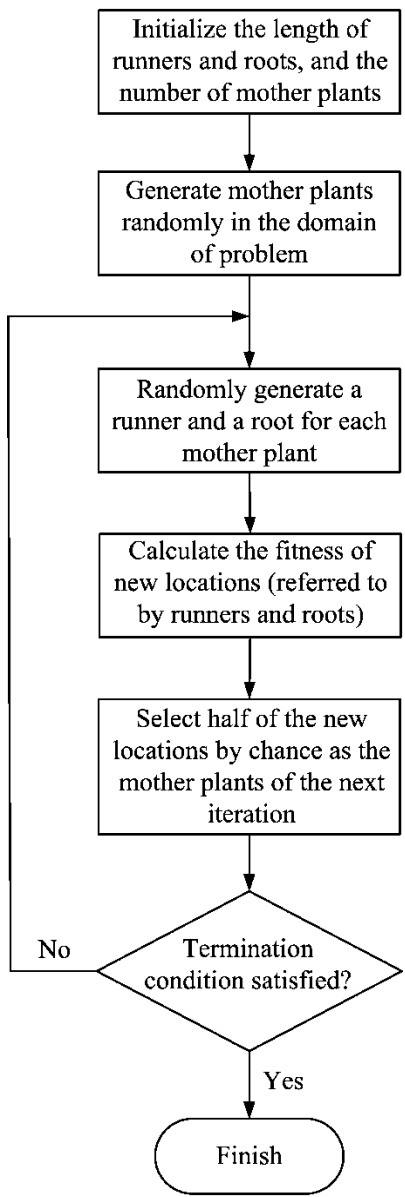

Fig. 2 Flowchart of the strawberry algorithm.

## 3. Mathematical Explanation of the Strawberry Algorithm

In this section we discuss on finding the solution of the following unconstrained optimization problem

$$\min f(\mathbf{x}), \qquad \mathbf{x}_l \leq \mathbf{x} \leq \mathbf{x}_u, \qquad (1)$$

where $f : \mathbb{R}^m \to \mathbb{R}$ is the $m$–variable objective (cost) function to be minimized, $\mathbf{x} \in \mathbb{R}^m$ is the solution vector to be calculated, and $\mathbf{x}_l, \mathbf{x}_u \in \mathbb{R}^m$ are two vectors indicating the lower and upper bounds of the variables.

According to the previous discussion, in order to solve the above problem first we randomly generate $N$ points in the domain of problem, each called a *mother plant*. Then, at each iteration each mother plant generates two random points: one very close to itself and another one faraway from it. The close and far points model the roots and runners (daughter plants) in nature, respectively. The roots are used to search around the locations of mother plants and the runners are used to search the locations considerably farther from them. Hence, the runners play the very important role of jumping over the local minimums, which effectively helps the algorithm to avoid trapping in these points (the function of runners in SBA is somehow similar to the *tunneling* property of particles in Quantum Annealing (QA) [24]).

In SBA, assuming that $\mathbf{x}_j(i) \in \mathbb{R}^m$ stands for the location of the *j*-th mother plant ($j = 1, \ldots, N$) at the *i*-th iteration, the matrix containing the locations of the corresponding runners and roots at this iteration, $\mathbf{X}_{prop}(i)$, is calculated as follows:

$$\mathbf{X}_{prop}(i) = [\mathbf{X}_{root}(i) \ \mathbf{X}_{runner}(i)] = [\mathbf{X}(i) \ \mathbf{X}(i)] + [d_{root}\mathbf{r}_1 \ d_{runner}\mathbf{r}_2], \qquad (2)$$

where $\mathbf{X}(i) = [\mathbf{x}_1(i) \ \mathbf{x}_2(i) \ \ldots \ \mathbf{x}_N(i)]$, $\mathbf{X}_{prop}(i) = [\mathbf{x}_{1,prop}(i) \ \mathbf{x}_{2,prop}(i) \ \ldots \ \mathbf{x}_{2N,prop}(i)]$, $\mathbf{X}_{root}(i) \in \mathbb{R}^{m \times N}$ and $\mathbf{X}_{runner}(i) \in \mathbb{R}^{m \times N}$ are matrices containing the locations of roots and runners as the following

$$\mathbf{X}_{root}(i) = [\mathbf{x}_{1,root}(i) \ \mathbf{x}_{2,root}(i) \ \ldots \ \mathbf{x}_{N,root}(i)], \quad (3)$$

$$\mathbf{X}_{runner}(i) = [\mathbf{x}_{1,runner}(i) \ \mathbf{x}_{2,runner}(i) \ \ldots \ \mathbf{x}_{N,runner}(i)], \quad (4)$$

$\mathbf{r}_1, \mathbf{r}_2 \in \mathbb{R}^{m \times N}$ are random matrices whose entries are independent random numbers with uniform distribution (or any other distribution) in the range $[-0.5, 0.5]$ (of course, the range is arbitrary), $d_{root}$ and $d_{runner}$ are two scalars representing the distance of roots and runners from the mother plant, respectively (often we have $d_{runner} \gg d_{root}$), and $N$ is the number of mother plants.

Note that according to (2) two points are generated for every mother plant: one very close to it (i.e., the root) and another one considerably faraway from it (i.e., the runner). Hence, application of (2) duplicates the number of computational agents, that is, $\mathbf{X}_{prop}(i)$ has $2N$ columns (potential solutions to the problem) while $\mathbf{X}(i)$ had only $N$ columns.

After calculation of $\mathbf{X}_{prop}(i)$, we use some sort of selection scheme (such as roulette wheel) to select $N$ columns (among the $2N$ columns) of $\mathbf{X}_{prop}(i)$ based on their performance such that *better* vectors have a higher chance to be selected (the selected vectors will be considered as the mother plants of the next iteration). In practice it is observed that a combination of elite and random selection leads to the best results. More precisely, in the proposed SBA half of the required $N$ mother plants are selected by elite-selection (which are exactly equal to the best vectors of $\mathbf{X}_{prop}(i)$) and the other half are selected by roulette wheel among the columns of

$\mathbf{X}_{prop}(i)$. Before applying roulette wheel, first the fitness of the $j$th column of $\mathbf{X}_{prop}(i)$, denoted as $fit(\mathbf{x}_{j,prop}(i))$, is calculated through the following equation (which is similar to ABC):

$$fit(\mathbf{x}_{j,prop}(i)) = \begin{cases} \dfrac{1}{a + f(\mathbf{x}_{j,prop}(i))} & f(\mathbf{x}_{j,prop}(i)) > 0 \\ a + \left| f(\mathbf{x}_{j,prop}(i)) \right| & f(\mathbf{x}_{j,prop}(i)) \leq 0 \end{cases}, \quad j = 1, \ldots, N, \quad (5)$$

where $f(.)$ is the cost function to be minimized and $a \geq 0$ is a parameter considered equal to zero in the simulations of this paper (this parameter can be used to adjust the selectivity property of the roulette wheel). After calculation of the fitness values, the probability of choosing the $j$th column, $p_j$, is calculated as follows

$$p_j = \frac{fit(\mathbf{x}_{j,prop}(i))}{\sum_{k=1}^{N} fit(\mathbf{x}_{k,prop}(i))}. \quad (6)$$

The selected vectors will be considered as the mother plants of the next iteration. Note that many other methods can also be used instead of (5) for calculating the fitness of solution vectors (see [25] for a comprehensive survey of fitness approximation in evolutionary computation). For example, in the so-called *ranking* method (often used in GA) first we sort the solution vectors in descendent order such that the worst and best vectors stand at positions 1 and *N*, respectively. Then, the fitness of the solution standing at position $pos$, $fit(pos)$, is calculated as follows

$$fit(pos) = 2 - SP + 2(SP - 1) \times \frac{pos - 1}{N - 1}, \quad (7)$$

where $SP \in [1, 2]$ is an arbitrarily chosen constant ($SP = 2$ leads to the most possible difference between the fitness of solutions, while $SP = 1$ makes no difference between the worst and best solutions).

Figure 3 shows the above mentioned procedure for $N = 3$ in a typical simulation. In this picture the black, red, and blue arrows and points refer to the first, second, and third iterations, respectively. First consider the black points and arrows (the first iteration). The points denoted as $\mathbf{x}_1(0)$, $\mathbf{x}_2(0)$ and $\mathbf{x}_3(0)$ are the initial data points generated randomly in the domain of problem. As it is observed, two arrows are connected to each initial random point: one with the *root* index in the vicinity of the corresponding point and the other with *runner* index in a farther location from it. These arrows show that any initial random point generates two points, which are located at the end of the corresponding arrows with *root* and *runner* indices (here, totally 6 new points are generated). The elite-selection and roulette-wheel select 3 out of these 6 points based on pure talent and chance, respectively (the arrows which refer to the points not selected are shown by dashed line and the corresponding points are omitted). The selected points, which are shown by red circles, are the points that will be used in the next iteration for further search. As it can be observed in this figure, the runners of $\mathbf{x}_1(0)$ and $\mathbf{x}_2(0)$, and the root of $\mathbf{x}_3(0)$ are selected as the initial points to be used in Iteration 1. These points are denoted as $\mathbf{x}_1(1)$, $\mathbf{x}_2(1)$ and $\mathbf{x}_3(1)$. Again, in Iteration 1 two arrows and points are generated for each of the selected points of previous iteration, and 3 out of them are selected by using the selection scheme. The selected points are shown by blue circles. This procedure is repeated again until a predetermined termination condition is satisfied. Note that the indices of points in Fig. 3 may be subjected to changes during iterations. For example, it is observed that in the second iteration neither the root nor the runner of $\mathbf{x}_1(1)$ is selected by roulette wheel or elite selection, and consequently, $\mathbf{x}_{1,root}(2)$ is not generated from $\mathbf{x}_1(1)$. In fact, at each iteration of SBA the algorithm may decide to continue the search either around the roots, or runners, or a combination of the roots and runners of the previous iteration. Small and large circles in Fig. 3 represent the radius of local and global searches, respectively. The pseudocode of SBA is presented in Table 1. Note that when the scale of variables is very different, position of mother plants can be calculated as the following

$$\mathbf{X}_{prop}(i) = [\mathbf{X}_{root}(i) \ \mathbf{X}_{runner}(i)] = [\mathbf{X}(i) \ \mathbf{X}(i)] + [\mathbf{d}_{root} \odot \mathbf{r}_1 \ \mathbf{d}_{runner} \odot \mathbf{r}_2], \qquad (8)$$

where $\mathbf{d}_{root}, \mathbf{d}_{runner} \in \mathbb{R}^m$ are the deterministic vectors representing the length of runners and roots at each dimension, $\odot$ is the Hadamard (element by element) matrix product operator, and other parameters are defined as before.

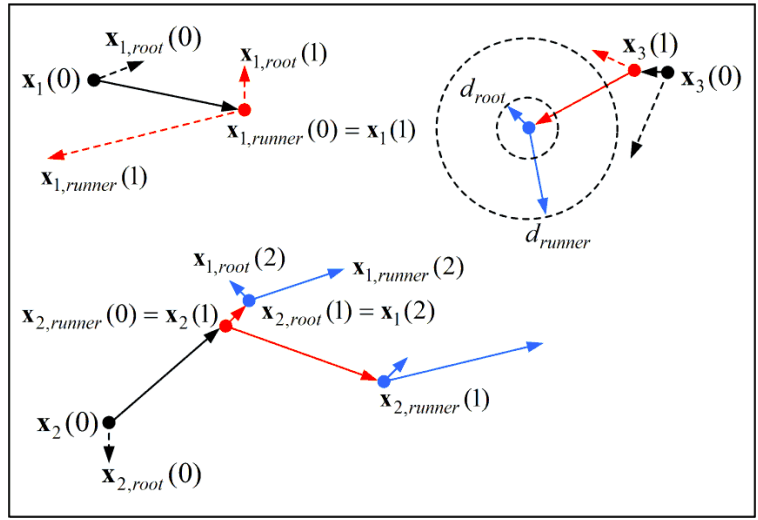

Fig. 3 Schematic explanation of the strawberry algorithm. The algorithm develops both runners (long arrows) and roots (small arrows) at every iteration for global and local search, respectively.

Table 1. The pseudocode of SBA.

---

Initialize $d_{root}$, $d_{runner}$, and $N$.

Set $i:=0$

Randomly initialize $N$ vectors (mother plants) in the domain of problem. Consider each of these random vectors as a column of $\mathbf{X}(0)$.

**WHILE** (the termination conditions are not met)

**Duplication:** For each vector (mother plant) calculate a runner (daughter plant) and a root from (2) and consider each of the resulted vectors as a column of $\mathbf{X}_{prop}(i)$.

**Elimination:** Calculate the fitness of runners and roots (i.e., the fitness of columns of $\mathbf{X}_{prop}(i)$) from (5), and then use elite selection and roulette wheel to select $N$ vectors (mother plants of the next iteration) among these $2N$ runners and roots. The selected vectors are considered as the columns of $\mathbf{X}(i+1)$.

Set $i:=i+1$

**END WHILE**

---

Here, it should be emphasized that unlike most of the other algorithms, in SBA the agents do not share any information about their solutions and the duplication-elimination procedure guarantees the movement of particles toward the optimum point. Of course, the lack of communication between agents reduces the computational cost of algorithm; however, $2N$ function evaluations are needed at each iteration. Another advantage of the proposed algorithm is that it has only three parameters to tune: $d_{root}$, $d_{runner}$, and $N$, which can often be selected by an easy trial and error. As a rule of thumb, $d_{root}$ should be selected sufficiently large such that the runners can jump over the hills. In fact, some algorithms (such as the standard PSO, ACO, ABC, and FA) have the property that all computational agents eventually converge to a single point (local or global optima), and after that the algorithm has a very small chance to escape from the local optima and find a better solution. On the other hand in of some the algorithms (such as GA and electromagnetism-like algorithm (EM)) the agents never converge to a single point, and consequently, continuing the iterations is always helpful for finding possible better solutions. SBA belongs to the second category.

### 3.1 Possible Modifications to the Proposed SBA

In this section we propose some possible modifications to the SBA presented in previous section, which may be used to generate, possibly more powerful, new explanations of this algorithm. These proposed modifications are not examined by the author and can be considered as the subjects of future studies.

The first point is that it is not necessary to keep the values of $d_{root}$ and $d_{runner}$ constant during the simulation; that is, the radius of local and global searches can be subjected to changes. As a general fact, most of the algorithms begin their job with a kind of global search and then, by increasing the number of iterations, they are intended to do more and more local searches. For example, in PSO the speed of particles is monotonically decreased by increasing the number of iterations (the particles with high and low speeds are suitable for global and local searches, respectively). Similarly, in GA (as it is implemented in the optimization toolbox of Matlab), amplitude of the changes caused by mutation is decreased by increasing the number of iterations. The proposed SBA performs both the local and global searches simultaneously from the beginning to end with two constant radiuses. This kind of search is the advantage of SBA, however, more effective algorithms may be obtained by varying the radius of these two searches in an appropriate manner. For example, in nature it is observed that a mother plant located in a very good place generates more runners and roots compared to another one that is not located in such a good place (in nature each strawberry mother plant may generate up to 30-40 runners). It is also observed that a plant which is located in a good place is more probable to generate longer runners than usual. These observations motivate us to modify SBA such that the mother plants with larger fitness values generate more runners and roots, or equivalently, the length of runners and roots can be increased after certain number of successive iterations in which no improvement is occurred.

Another possible modification to SBA is that we can perform the duplication-elimination procedure with a multiple rather than two (regardless of the fitness value of agents). It means that each mother plant can generate $n$ runners and $n$ roots (instead of one), and then the algorithm selects $N$ mother plants among them.

## 4. Application of SBA to Classical Test Functions

In the following, we apply SBA to find the minimum of two classical test functions of different dimensions, and briefly compare the results with those obtained by using PSO and GA. It should be emphasized that the aim of these comparisons is not to conclude that SBA works better than other algorithms, and these are used just to convince the reader about the applicability of SBA to real world problems. Detailed comparison of different algorithms is a tricky task which is not discussed in this paper and can be found in the literature (see, for example [26]). The Matlab

codes of the following simulations, as well as many others, can be downloaded from

https://drive.google.com/file/d/0Bwl6hPor5o61OElmMWhtbEJadkU/edit?usp=sharing .

**Example 1.** The *n*-dimensional Rastrigin function is defined through the following equation

$$f(\mathbf{x}) = 10n + \sum_{i=1}^{n}\left[x_i^2 - 10\cos(2\pi x_i)\right], \tag{9}$$

where $x_i \in [-5.12, 5.12]$ ($i = 1, 2, \ldots, n$). This function has a global minimum at $\mathbf{x} = 0$ for which we have $f(0) = 0$. The main reason for the difficulty of finding the global minimum of this function is that it is surrounded by many local minimums where each local minimum has a very small difference (from the objective function value point of view) with the other one in its vicinity.

Figure 4 shows the locations of the mother plants generated by SBA during 100 iterations in a typical simulation of the two dimensional Rastrigin function assuming $d_{root} = 0.2$, $d_{runner} = 2.5$, and $N = 5$ (note that many points are plotted in almost the same point in this figure). In this simulation the best solution is obtained as $(x_1^*, x_2^*) = (-9 \times 10^{-4}, -4.4 \times 10^{-3})$ which is fairly close to the global best solution. Figure 5 shows the minimum, maximum and mean values obtained for the two dimensional Rastrigin function using SBA versus iteration number (the results are averaged over 100 runs assuming $d_{root} = 0.2$, $d_{runner} = 2.5$, and $N = 5$). This figure clearly shows the very fast convergence of SBA.

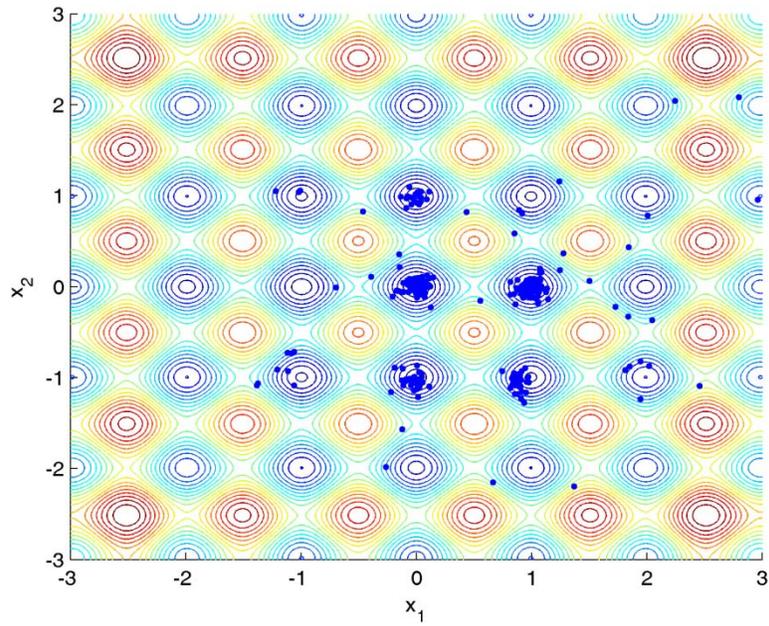

Fig. 4 Locations of mother plants in a simulation of two dimensional Rastrigin function.

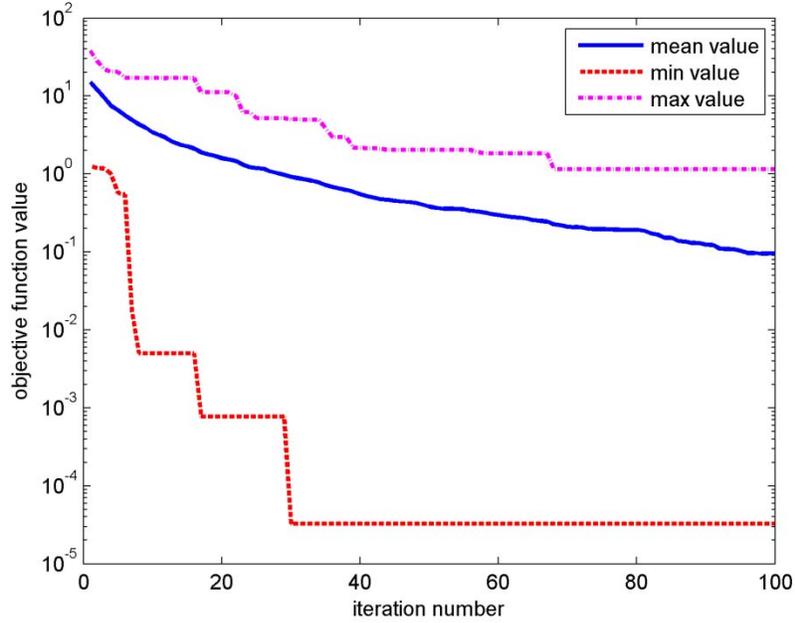

Fig. 5 Mean, minimum, and maximum of two dimensional Rastrigin function when SBA is applied (the results are averaged over 100 runs assuming $d_{root} = 0.2$, $d_{runner} = 2.5$, and $N = 5$).

Figure 6 compares SBA with the standard GA and PSO when they are applied to the four dimensional Rastregin function (the results are averaged over 100 runs assuming $d_{root} = 0.2$, $d_{runner} = 2.5$, and $N = 5$). Note that since the number of function evaluations in SBA is two times the number of mother plants, the number of individuals and particles in GA and PSO (in the simulation of Fig. 6) is considered equal to 10 to make sure that all of these algorithms apply the same number of function evaluations. Note also that we have applied SBA, GA and PSO assuming certain values for the parameters used in these algorithms and better results may be obtained by changing these values (however, we did our best to obtain the best results in each case). More precisely, in dealing with GA, 60%, 30%, and 10% of individuals are generated by crossover, reproduction, and mutation operators, respectively, and each variable is coded with a 10 bit binary string. In addition, in all cases the fitness of computational agents is calculated through (5) assuming $a = 0$. Finally, it should be noted that since the initial random population in GA is generated directly by using binary strings, in the following simulations the curves corresponding to GA begin from different initial points compared to SBA and PSO. However, as it is expected, the rate of decreasing the value of objective function is smaller when GA is applied.

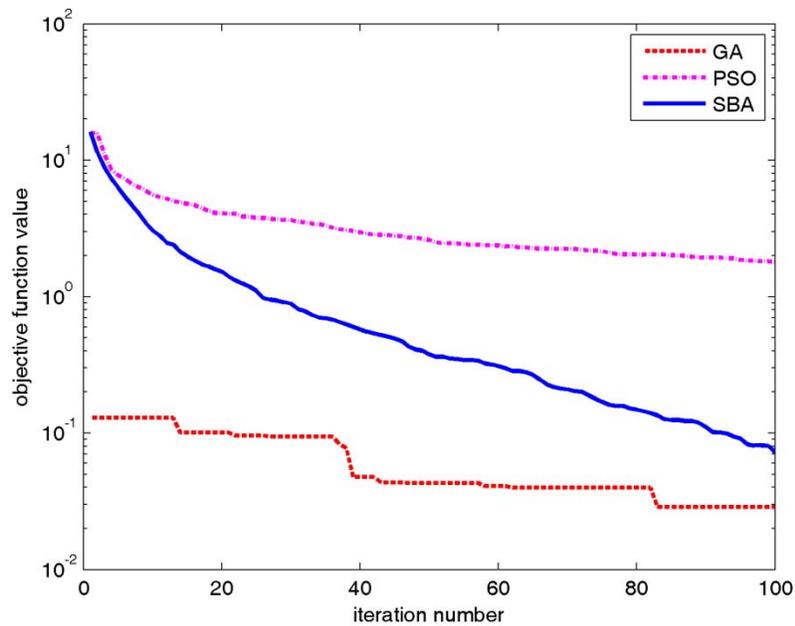

Fig. 6 Results of applying GA, PSO, and SBA to the four dimensional Rastrigin function.

Figure 7 shows the simulation results for the 20 dimensional Rastrigin function when GA, PSO, and SBA are applied (the results are averaged over 100 runs). In this figure, the number of individuals, particles, and mother plants in GA, PSO, and SBA is considered equal to 100, 100, and 50, respectively ($d_{root} = 0.2$ and $d_{runner} = 2.5$ is applied in SBA). The high performance of SBA can be explained through the effect of runners, which help to algorithm to jump over the hills.

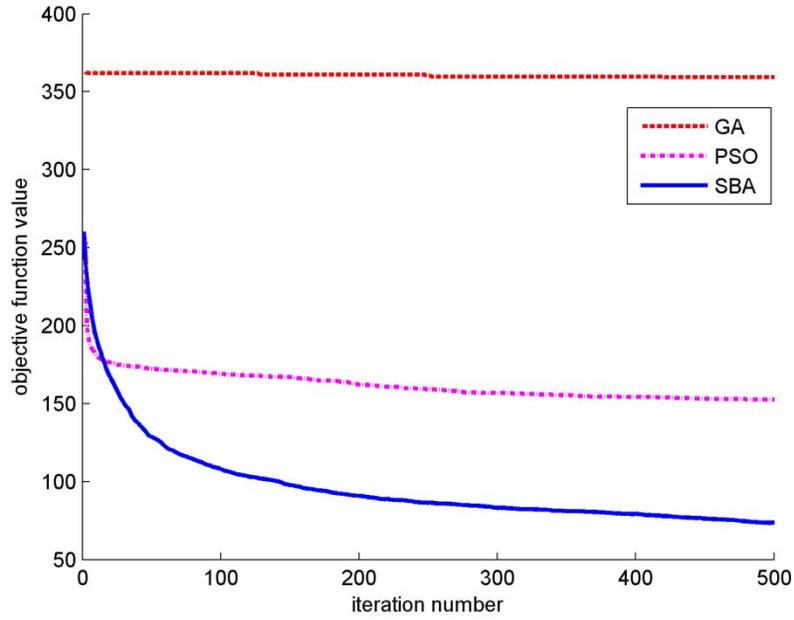

Fig. 7 Results of applying GA, PSO, and SBA to the 20 dimensional Rastrigin function.

**Example 2.** The $n$ variable Griewank function is defined as the following

$$f(\mathbf{x}) = \frac{1}{4000}\sum_{k=1}^{n} x_i^2 - \prod_{k=1}^{n} \cos\left(\frac{x_i}{\sqrt{k}}\right), \tag{10}$$

where $x_i \in [-600, 600]$ ($i = 1, 2, \ldots, n$). This function has a global minimum at $\mathbf{x} = 0$ for which we have $f(0) = 0$. As a well-known fact, the number of local minimums of this function is exponentially increased by increasing the value of $n$ (see [27] for more information about the number of local minimums of this function). According to the huge number of local minimums, which are very similar and close to each other, finding the global minimum of this function is a challenging task even for $n = 2$. Figure 8 shows the locations of mother plants in a typical simulation when SBA with $d_{root} = 10$, $d_{runner} = 400$, and $N = 5$ is applied to the two dimensional

Griewank problem (the simulation stops after 1000 iterations). Any blue point in this figure indicates the position of a mother plant in some iteration. As it can be observed in this figure, the overall trend of mother plants is toward the global minimum at origin. Note that each circle in Fig. 8 itself consists of a huge number of local minimums and the region between any two circles is also full of almost identical local minimums (that is why this region looks white in this plot). Figure 9 shows Fig. 8 when focused around the origin. This figure clearly shows the tremendous number of local minimums around the global minimum at the origin and also the concentration of mother plants around this point. Here it is worth to emphasize that at the beginning of SBA the mother plants pass through the local minimums mainly by means of their runners and when they became closer to the origin they do this task mainly through their roots.

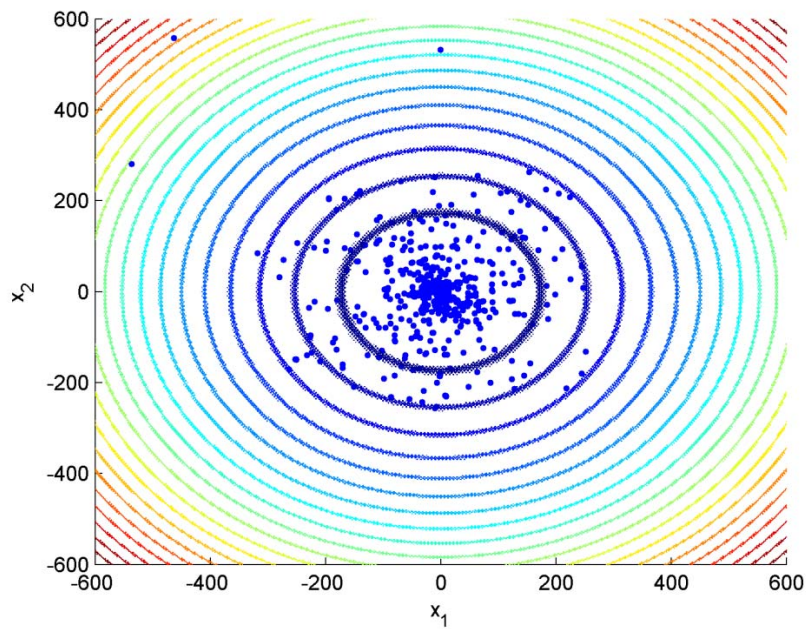

Fig. 8 Positions of mother plants in the simulation of two dimensional Griewank function. The mother plants jump over the local minimums faraway from the origin through their runners.

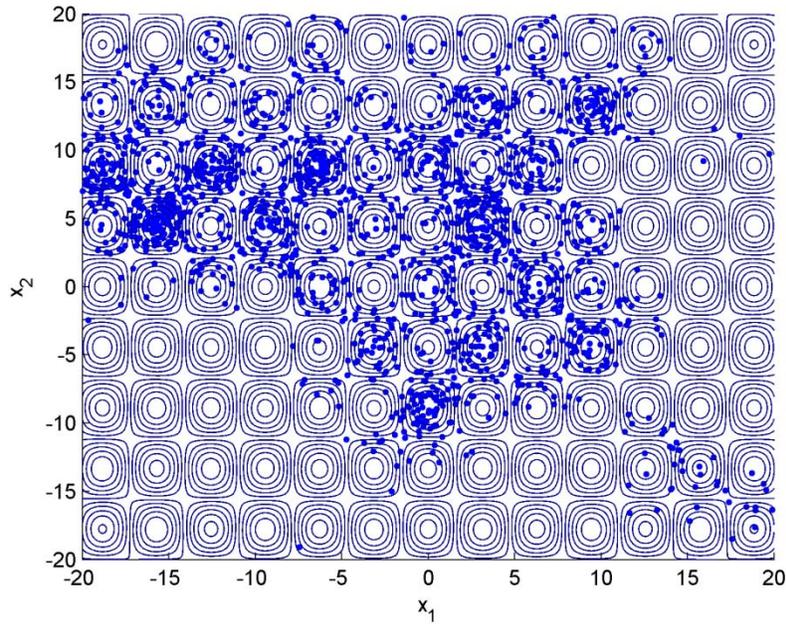

Fig. 9 Figure 8 focused around the origin. The mother plants jump over the local minimums around the origin through their roots.

Figure 10 shows the mean, maximum and minimum value of the two dimensional Griewank function versus iteration number when SBA with $d_{root} = 10$, $d_{runner} = 400$, and $N = 5$ is applied (the results are averaged over 100 runs). Figure 11 shows the average value of this function versus the iteration number when GA, PSO, and SBA are applied (GA and PSO are applied using 10 computational agents and again the results are averaged over 100 runs).

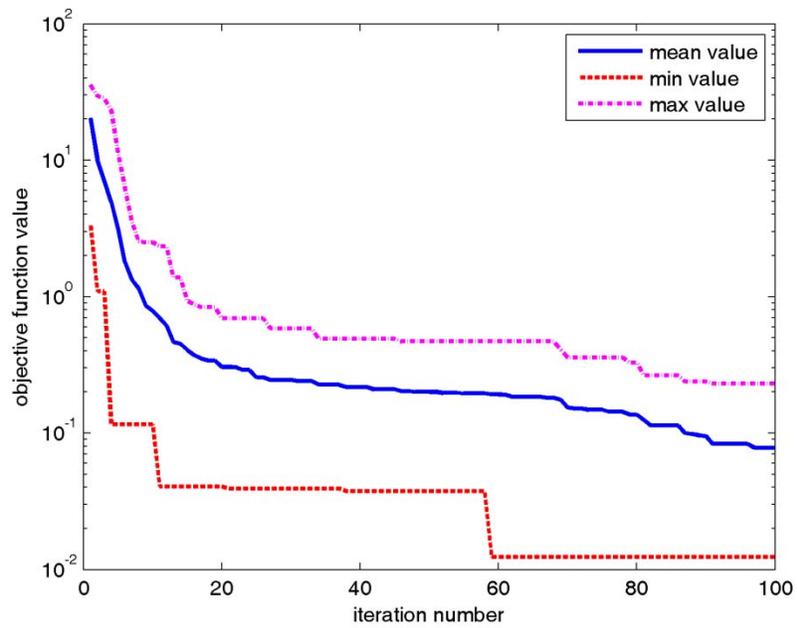

Fig. 10 Mean, minimum, and maximum of the two dimensional Griewank function when SBA is applied (the results are averaged over 100 runs assuming $d_{root} = 10$, $d_{runner} = 400$, and $N = 5$).

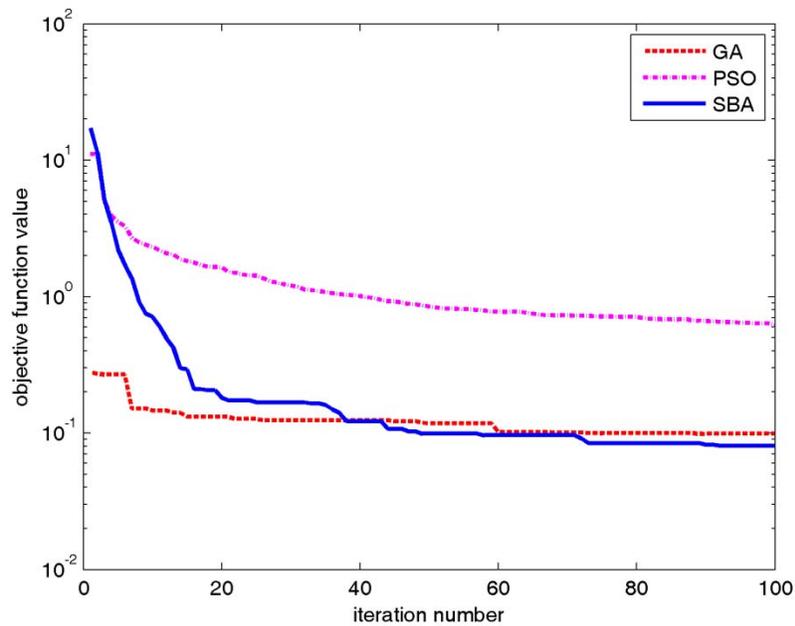

Fig. 11 Results of applying GA, PSO, and SBA to the two dimensional Griewank function (the results are averaged over 100 runs).

Figure 12 shows the results of applying GA, PSO, and SBA to the 20 dimensional Griewank function. The results are averaged over 100 iterations and the parameters used in algorithms are considered similar to the previous simulation. This figure clearly shows the high performance of SBA in dealing with this high dimensional problem.

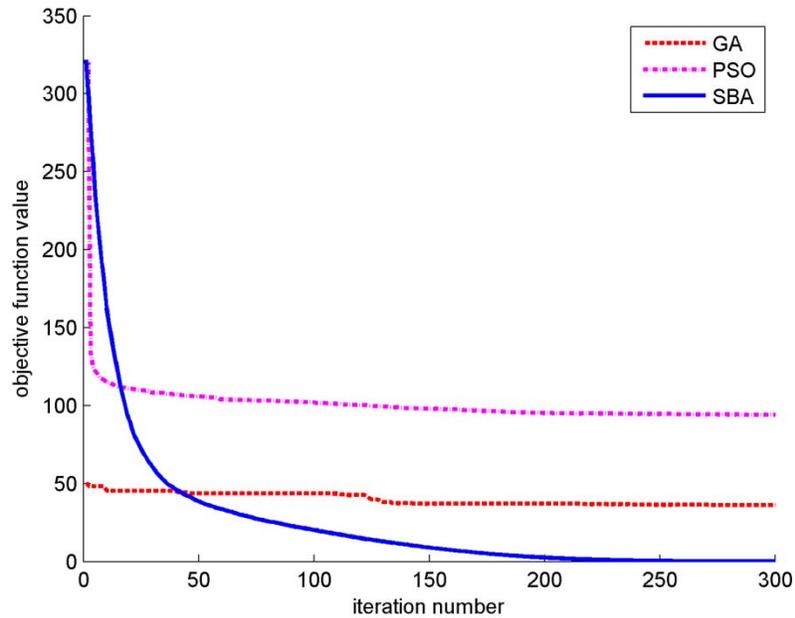

Fig. 12 Results of applying GA, PSO, and SBA to the 20 dimensional Griewank function (the results are averaged over 100 runs).

## 5. Application of SBA for Solving an Open Problem in the Field of Robust Control Theory

In this example we introduce an unsolved problem in the field of robust control theory and provide a numerical solution for it using SBA. Consider the single-input single-output (SISO) feedback system shown in Fig. 13 where the uncertainty of process is modeled with the full multiplicative input uncertainty ($\|\Delta_I\|_\infty \leq 1$) [28]. In this figure assuming that the process nominal transfer function, $G(s)$, and uncertainty weight, $w_I(s)$, are stable and known, the controller which simultaneously satisfies nominal performance (NP), robust stability (RS), and robust performance (RP) is obtained by solving the following $H_\infty$-norm inequality [28]:

$$\left\| |w_P S| + |w_I T| \right\|_\infty < 1, \tag{11}$$

where $S = (1+KG)^{-1}$, $T = KG(1+KG)^{-1}$, $K$ is the unknown controller, and $w_P$ is the stable weight function used to determine the desired performance (i.e., it is assumed that the nominal feedback system with $\Delta_I = 0$ has a desired performance if and only if we have $|S| < |w_P|^{-1}$).

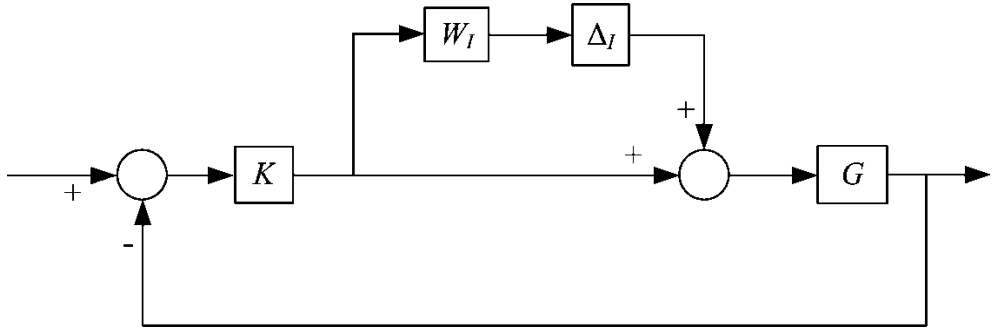

Fig. 13 A feedback system with multiplicative uncertainty at the process input.

Note that the only unknown parameter in (11) is the transfer function of controller, $K$. As a classical result [28], the controller which satisfies (11) also guarantees NP, RS, and RP, but the nominal stability (NS) is not guaranteed in this case, i.e., the controller obtained by solving (11) may result in an unstable feedback system. Hence, any solution for (11) must also be checked from the NS point of view. In the following, we will consider the NS condition as a constraint for a minimization problem.

At this time there is no method available to find a $K$ which simultaneously satisfies (11) and guarantees the stability of nominal feedback system [28]. Instead, algorithms like *hinfsyn* in Matlab, find $K$ such that the following closely related mixed sensitivity $H_\infty$ condition is satisfied

$$\left\| \begin{matrix} w_p S \\ w_I T \end{matrix} \right\|_\infty = \max_\omega \sqrt{|w_p S|^2 + |w_I T|^2} < 1. \tag{12}$$

Although the above condition is within a factor of at most $\sqrt{2}$ to condition (11), they are not representing exactly the same conditions [28]. Note that similar to (11), the NS is not included in (12). The very important point in relation to (12) (as well as (11)) is that it may or may not have a solution. However, when it has, often it is not unique (infinity many solutions exist in general). For that reason, this problem is often formulated as a minimization problem in which

$\gamma \triangleq \| |w_P S| + |w_I T| \|_\infty$ (or $\gamma \triangleq \left\| \begin{matrix} w_p S \\ w_I T \end{matrix} \right\|_\infty$) is minimized by suitable choice of the (stabilizing) $K$.

According to the above discussion it is obvious that the robust control problem under consideration *really* has a solution if for some stabilizing $K$ we have $\gamma = \| |w_P S| + |w_I T| \|_\infty < 1$.

The aim of this example is to calculate $K$ directly from (12). For this purpose we find $K$ by solving the following constrained minimization problem:

$$\begin{aligned} \min_K \gamma &= \| |w_P S| + |w_I T| \|_\infty \\ \text{s.t. } & 1 + KG \neq 0, \operatorname{Re}\{s\} \geq 0 \end{aligned}, \tag{13}$$

where the constraint $1 + KG \neq 0$, $\operatorname{Re}\{s\} \geq 0$ represents the stability of the nominal closed-loop system. A classical approach for solving constrained optimization problems is to express the problem as an equivalent one without constraint (a survey of techniques used for handing the constraints can be found in [29]). As the simplest method, the solution of (13) can be obtained by solving the following unconstrained optimization problem

$$\min F(\mathbf{x}) \triangleq \| |w_P S| + |w_I T| \|_\infty + \lambda \langle g(\mathbf{x}) \rangle, \tag{14}$$

where **x** is a vector containing the unknown parameters of controller, $\lambda > 0$ is the penalty factor, $g(\mathbf{x})$ is equal to the real part of the rightmost pole of the nominal closed-loop system (i.e., the system of Fig. 13 when $\Delta_I = 0$), and $\langle . \rangle$ is the bracket function defined as the following

$$\langle g(\mathbf{x}) \rangle \triangleq \begin{cases} g(\mathbf{x}) & g(\mathbf{x}) \geq 0 \\ 0 & g(\mathbf{x}) < 0 \end{cases}. \tag{15}$$

Note that according to (14) and (15), when the candidate solution **x** does not violate the stability condition, the real part of the rightmost pole of the nominal closed-loop system is negative, and consequently, the cost function is simply equal to $F(\mathbf{x}) = \| |w_P S| + |w_I T| \|_\infty$. But, when the stability condition is violated, the second term in the right hand side of (14) puts an extra force on the algorithm to obtain solutions that lead to stable feedback systems. Clearly, larger the value of $\lambda$, higher the force on algorithm to find the solutions that stabilize the feedback system.

In the following we apply the above procedure to find a controller for a process with nominal transfer function $G(s) = 1000/(0.1s^2 + s)$ assuming $w_I = 0.2$ and $w_P = (0.2s+1)/(s+0.001)$. In this example $G(s)$ is the approximate model of a DC motor and $w_I = 0.2$ represents the fact that this model has 20% uncertainty at each frequency. Considering the fact that the order of $K$ is at least equal to the order of process plus the weights $w_I$ and $w_P$ [29], the transfer function of controller is considered as the following

$$K(s) = \frac{b_2 s^2 + b_1 s + b_0}{s^3 + a_2 s^2 + a_1 s + a_0}, \tag{16}$$

where $a_i, b_i \in \mathbb{R}$ ($i = 0,1,2$) are unknown parameters of the controller to be determined by solving (14) (note that here we have $\mathbf{x} = [a_0\ a_1\ a_2\ b_0\ b_1\ b_2]$). Assuming $\lambda = 10^5$, $N = 50$,

$d_{runner} = 10^{10}$, $d_{root} = 10^8$, $a = 0$, and $\mathbf{x}_u = -\mathbf{x}_l = 10^{10}$, and after few runs the solution of (14) using SBA is obtained as

$$K(s) = \frac{9.8371 \times 10^8 s^2 + 5.5818 \times 10^9 s + 1.9011 \times 10^8}{s^3 + 2.2109 \times 10^4 s^2 + 5.0171 \times 10^{10} s - 8.6854 \times 10^8}, \quad (17)$$

for which we have $\gamma = \||w_P S| + |w_I T|\|_\infty = 0.2799$ and $g(\mathbf{x}^*) = -0.0343$ (since the solutions that lead to unstable feedback systems are quite useless in practice, the value of $\lambda$ is considered very large to make sure that such solutions are abandoned rapidly by SBA). Figure 14 shows $|w_P S| + |w_I T|$ versus frequency (in log scale) for the controller given in (17). As it is observed, the plot is under 0dB at all frequencies which guarantees NP, RS, and RP (the NS is achieved by the negative value obtained for $g(\mathbf{x}^*)$). Figure 15 shows the objective function value (as defined in (14)) versus the iteration number in this example (the results are averaged over 20 runs). It is concluded from this figure that the maximum value of objective function becomes smaller than unity in less than 50 iterations, which means that the algorithm is able of finding a solution for this problem in about 50 iterations. Figure 16 shows the unit step response of the closed-loop system under consideration for 30 different random $\Delta$'s. As it is observed, the closed-loop system with the controller given in (17) exhibits a very good robust performance.

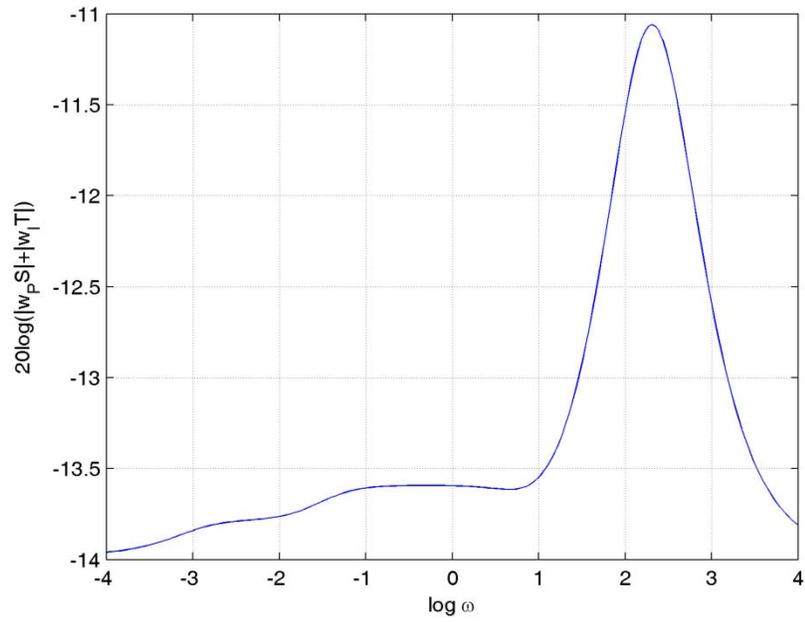

Fig. 14 Plot of $|w_P S| + |w_I T|$ versus frequency.

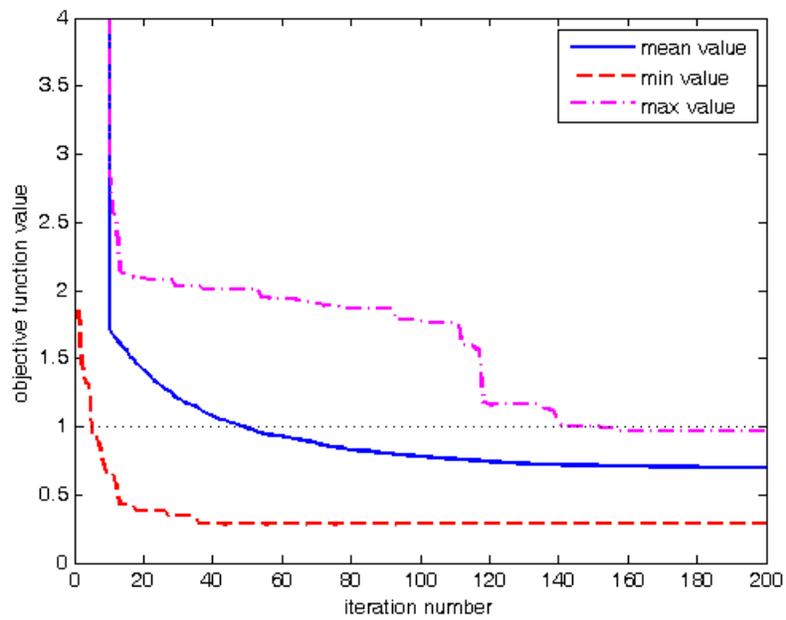

Fig. 15 Value of the cost function given in (14) versus iteration number.

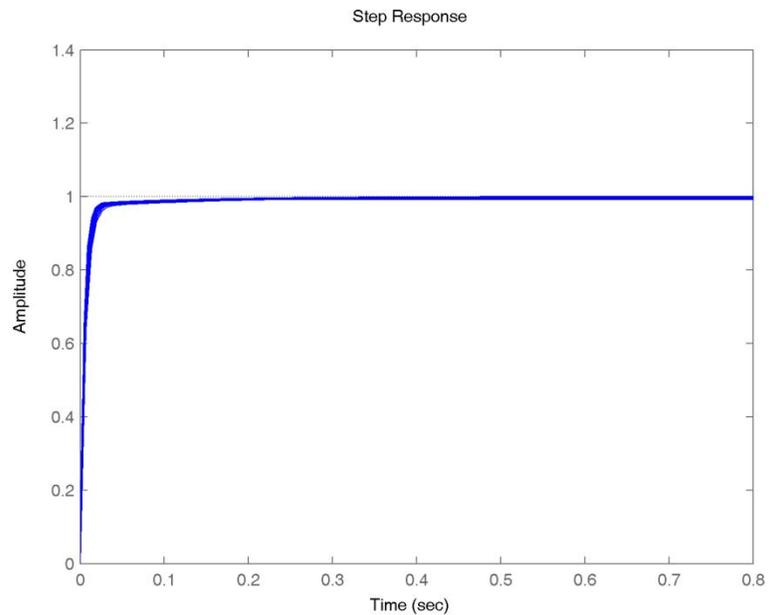

Fig. 16 Unit step response of the feedback system for 30 different random uncertainties.

## Conclusion

A numerical optimization algorithm inspired by the strawberry plant is presented in this paper. The proposed algorithm has the property that simultaneously applies local and global searches to find the global best solution. This algorithm is applied to two test functions and the results show that it is capable of finding the optimum point of non-convex cost functions with a high efficiency. Moreover, the proposed algorithm is successfully used to solve an open problem in the field of robust control theory.

# Acknowledgment

The author is indebted to F. Hojjati-Parast for pointing out the role of roots in plants as tools for local search for water resources and minerals.